\documentclass[a4paper,conference]{IEEEtran}
\ifCLASSINFOpdf
  \usepackage[pdftex]{graphicx}
   \graphicspath{{../pdf/}{../jpeg/}}
  \DeclareGraphicsExtensions{.pdf,.jpeg,.png}
\else
  \usepackage[dvips]{graphicx}
\fi
\usepackage{threeparttable}
\usepackage{hyperref}
\usepackage{amsmath}
\usepackage{amssymb}
\usepackage{algorithm}
\usepackage{algorithmic}

\usepackage{array}
\usepackage{multirow}
\ifCLASSOPTIONcompsoc
\usepackage[caption=false,font=normalsize,labelfont=sf,textfont=sf]{subfig}
\else
\usepackage[caption=false,font=footnotesize]{subfig}
\fi
\begin{document}
%
\title{Multi-scale Network with Attentional Multi-resolution Fusion for Point Cloud Semantic Segmentation}

\author{
\IEEEauthorblockN{Yuyan Li}
\IEEEauthorblockA{Department of Electrical Engineering \\and Computer Science\\
University of Missouri - Columbia\\
Email: yl235@umsystem.edu}
\and
\IEEEauthorblockN{Ye Duan}
\IEEEauthorblockA{Department of Electrical Engineering \\and Computer Science\\
University of Missouri - Columbia\\
Email: duanye@umsystem.edu}
}


%


\maketitle

\begin{abstract}
In this paper, we present a comprehensive point cloud semantic segmentation network that aggregates both local and global multi-scale information. First, we propose an Angle Correlation Point Convolution (ACPConv) module to effectively learn the local shapes of points. Second, based upon ACPConv, we introduce a local multi-scale split (MSS) block that hierarchically connects features within one single block and gradually enlarges the receptive field which is beneficial for exploiting the local context. Third, inspired by HRNet which has excellent performance on 2D image vision tasks, we build an HRNet customized for point cloud to learn global multi-scale context. Lastly, we introduce a point-wise attention fusion approach that fuses multi-resolution predictions and further improves point cloud semantic segmentation performance. 
Our experimental results and ablations on several benchmark datasets show that our proposed method is effective and able to achieve state-of-the-art performances compared to existing methods. 
\end{abstract}


%
\IEEEpeerreviewmaketitle

\section{Introduction}
A growing interest in 3D data representation and analysis has been stimulated by the emergence of technologies, such as autonomous driving, robotic navigation, virtual reality, etc. The popularity of point clouds, as opposed to other 3D representations, has been rising due to the development of 3D sensors (such as LiDAR), which produce point clouds as raw data. As point clouds emerged, many 3D deep learning methods are explored and developed. However, how to effectively represent unordered and irregular point clouds remains a key challenge for researchers.

Many 3D point cloud deep learning approaches are inspired by the successful methods designed for image-related tasks. For example, multi-view images approaches \cite{boulch2018snapnet, kalogerakis20173d, su2015multi}, voxel-based approaches \cite{le2018pointgrid,tchapmi2017segcloud} are developed to handle point clouds like regular 2D/3D images. But converting to these representations inevitably increases computation cost and leads to information loss. Therefore, recent approaches favor processing point clouds directly.  A point-based pioneer work, PointNet\cite{qi2017pointnet}, extracts local point-wise features using multi-layer perceptron (MLP) and combines them with global information through pooling. Despite its efficiency, PointNet does not aggregate local information. The follow-up work PointNet++ \cite{qi2017pointnet++} proposed a multi-scale set attraction module to learn neighborhood features as well as a hierarchical architecture. After PointNet++, many works \cite{hu2020randla, xu2021paconv, thomas2019kpconv, li2021spnet} have proposed more powerful local feature learning modules to capture the fine details of point cloud shapes. Several works also \cite{huang2019multi, zhang2019shellnet,li2020multi} introduced local multi-scale feature encoding operators.
Nevertheless, these studies mainly explore local feature encoding operators, global context learning is not well explored. 

Inspired by the previous works, we are aiming to develop a framework for extracting multi-scale local and global features that will be robust against point cloud density variability and able to encode both fine object details and global structural information. Locally, we propose an Angle Correlated Point Convolution operator, namely ACPConv (see Figure \ref{fig:kernel}) that aggregates neighborhood features using manually designed kernel points. Using ACPConv as a major component, we design a multi-scale split (MSS) block. MSS block first split input feature map into multiple subgroups of features, each of which is passed through an ACPConv, the output of ACPConv is split again, the first half is copied and used as output, the second half is concatenated with next subgroup feature (see Figure \ref{fig:hsblock}). The receptive field within the MSS block gradually increases through the hierarchical connection, demonstrating strong multi-scale representation ability. Globally, we leverage High-resolution Network (HRNet) \cite{SunXLW19,WangSCJDZLMTWLX19} which achieves state-of-the-art performances in image object detection, pose estimation, and semantic segmentation tasks.  We simplify HRNet (see Figure \ref{fig:hrnet}) and customize it for point cloud semantic segmentation. By connecting high-to-low resolution convolutions in parallel, HRNet maintains high-resolution representations and produces robust high-resolution representations by repeatedly fusing parallel convolutions. Furthermore, we observe that some semantic classes (i.e., painting, board) are better handled at high point cloud resolution, while some classes (i.e., wall, column) are more precisely handled at low resolution. Based upon this, we propose a multi-resolution attention fusion mechanism that learns to adaptively combine different resolution predictions. Specifically, we use various downsampling rates to control the resolution of point clouds. For each resolution, we predict point-wise per-class attention scores for re-weighting the predictions.  Our contributions can be summarized as follows:
\begin{itemize}
    \item We present a multi-scale framework that captures local fine details and global contextual information for point cloud semantic segmentation.
    \item We propose an attention mechanism that learns how to fuse multi-resolution point cloud predictions.
    \item Our experiments show that we achieve state-of-the-art performances on several point cloud benchmarks.
\end{itemize}

\section{Related Work}
\subsection{Point Cloud Learning Networks}
Recent years have seen the emergence of numerous point cloud learning techniques. To leverage the success of approaches designed for 2D images, many works \cite{boulch2018snapnet, kalogerakis20173d, su2015multi} proposed to use multi-view image representation. But these methods suffer from inherent geometric information loss. VMVF \cite{kundu2020virtual} proposed a novel synthetic rendered view-based method that overcomes the limitations and achieved great performances. Voxel-based methods \cite{le2018pointgrid,tchapmi2017segcloud} used volumetric grid representation which can be processed by 3D CNN. Voxel-based methods usually lead to expensive computation costs as well as information loss during voxelization. Recently, sparse convolution methods \cite{choy20194d,graham20183d} which maintain sparsity by computing only on active pixels have been proposed to greatly improve performance and efficiency. To overcome these limitations, researchers have been seeking approaches to directly process raw point clouds. PointNet \cite{qi2017pointnet} combines point-wise features extracted by MLPs with global pooling features. PointNet does not aggregate local features and therefore has limited performance. Later, PointNet++ \cite{qi2017pointnet} proposed a multi-scale local feature encoding module along with a hierarchical structure that enlarges the receptive field with point down/upsampling. 
Inspired by PointNet++, many recent works \cite{thomas2019kpconv,li2021spnet,li2021spnet,hu2020randla,boulch2020fkaconv,lei2020spherical} introduced more complex local encoding modules. For example, RandLa-Net \cite{hu2020randla} incorporates local spatial encoding and attentive pooling to preserve the local structure. KPConv \cite{thomas2019kpconv} mimics 2D image convolution using kernel points. The convolution weights of KPConv are determined by the Euclidean distances to kernel points. The positions of the kernel points are formulated as an optimization problem of the best coverage in a sphere space. Several attention-based aggregation methods \cite{wang2019graph,xie2018attentional} are proposed to model long-range dependencies and to re-weight features to focus on relevant information.
In our work, we choose to use a kernel point convolution ACPConv for local feature encoding. Combining ACPConv with multi-scale blocks effectively enlarges the receptive field and enables complex shape patterns to be learned.

\subsection{Multi-scale Representations}
Researchers have made significant progress on developing multi-scale methods for 2D vision tasks. Res2Net \cite{gao2019res2net} proposed a building block of CNN which represents multi-scale features at a granular level and increases the range of receptive fields for each network layer. HS-ResNet \cite{yuan2020hs} introduced a novel hierarchical split block to generate multi-scale feature representations. With regard to global multi-scale representation, PSPNet \cite{zhao2017pyramid} used a spatial pyramid pooling (SPP) module is used to assemble features at multiple scales based on the final layer of network trunk using a sequence of pooling and convolution operations. DeepLab \cite{chen2017deeplab} proposed Atrous Spatial Pyramid Pooling (ASPP) which employs atrous convolutions with different levels of dilation. More recently, HRNet \cite{SunXLW19,WangSCJDZLMTWLX19} achieves state-of-the-art performances for human pose estimation and semantic segmentation. HRNet maintains high-resolution through the entire network, and hierarchically connects high-to-low resolution convolution streams in parallel, repeatedly fuses the multi-scale features. In \cite{tao2020hierarchical}, an attention mechanism is introduced to predict how to adaptively combine multi-scale predictions together at a pixel level. 

Multi-scale information is crucial not only for 2D vision but also for the learning of point clouds. Many local multi-scale operators \cite{zhang2019shellnet, huang2019multi,li2020multi, li2021spnet} have been proposed. For example, ShellNet \cite{zhang2019shellnet} introduced concentric spherical shells to define multi-scale representative features. SPNet \cite{li2021spnet} used multiple groups of kernel point convolution. Moreover, some works proposed global multi-scale architectures for point cloud classification \cite{wang2018msnet}, and semantic segmentation \cite{qin2019pointdan,liang2019mhnet} tasks. Motivated by the previous works, we propose a comprehensive framework that extracts multi-scale features of point clouds. 
Our framework is able to capture both fine details and global structures with state-of-the-art performance due to multi-scale information.

\section{Method}
In this section, we will first describe the details of Angle Correlation Point Convolution and the local multi-scale split block which includes ACPConv as the major module. Then we will elaborate on the global multi-scale architecture and the multi-resolution network with attentional fusion.

\subsection{Angle Correlation Point Convolution (ACPConv)}
\label{sec:acpconv}

\begin{figure*}[t]
    \centering
    \includegraphics[width=16cm]{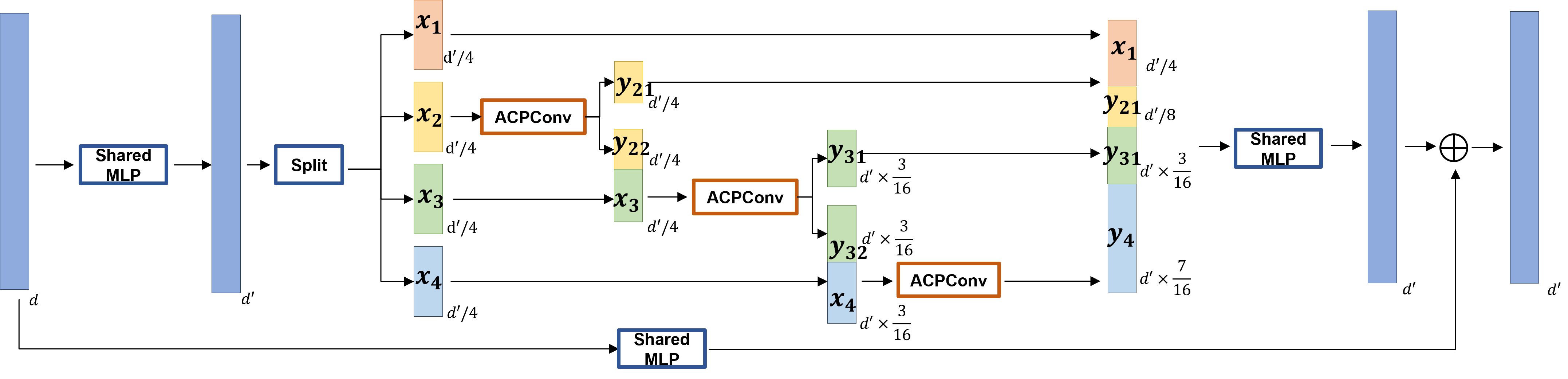}
    \caption{Illustration of our local multi-scale split (MSS) block. The feature maps are repeatedly split and hierarchically concatenated to gradually expand the receptive field and provide multi-scale representation capability.}
    \label{fig:hsblock}
\end{figure*}

\begin{figure}[tbh!]
    \centering
    \includegraphics[width=7cm]{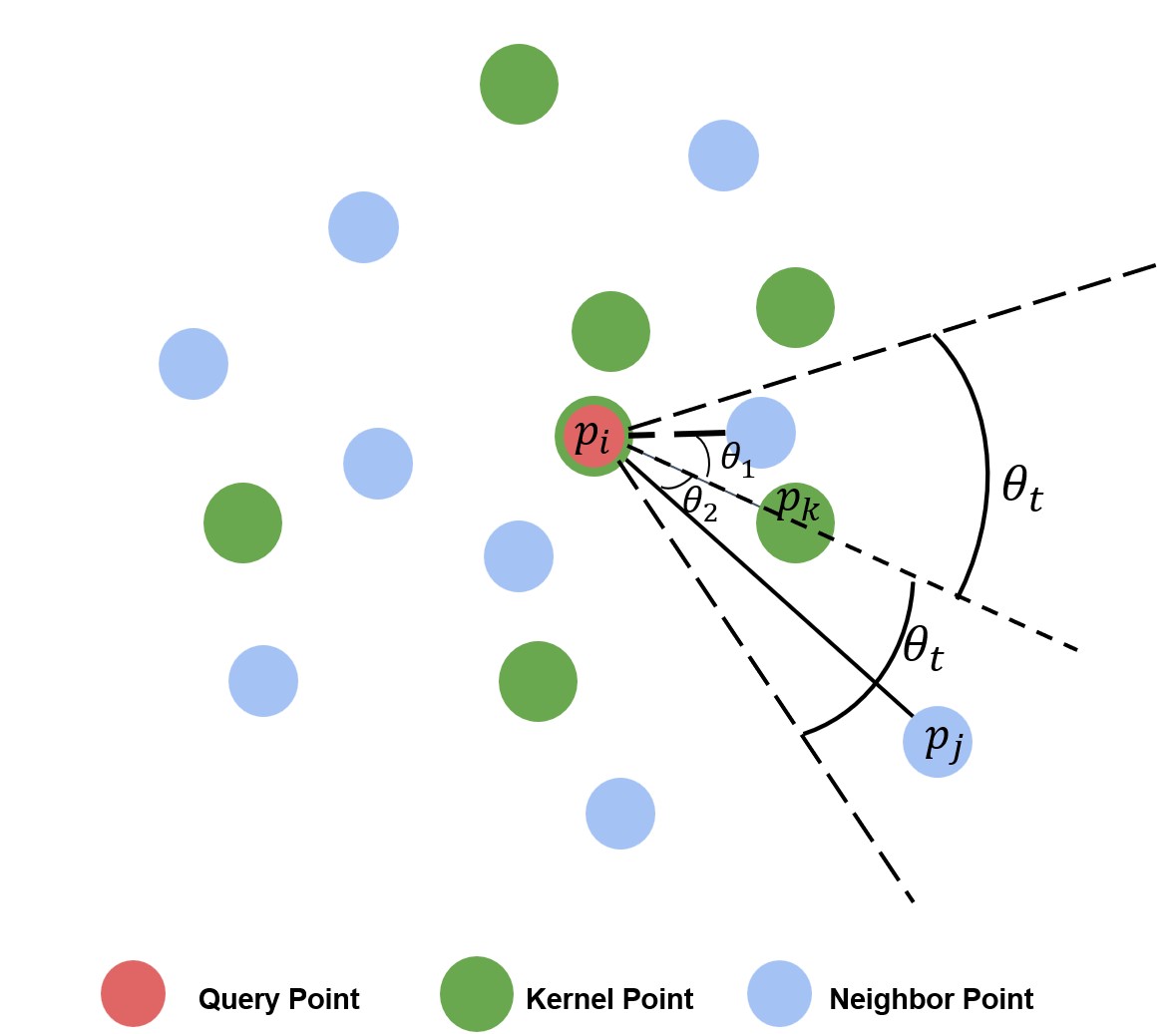}
    \caption{ACPConv illustration. For a query point (red), we use KNN search to locate $M$ neighbor points (blue). We randomly initialize $K$ kernel points (green) with one kernel occupying the center. For each kernel point, our proposed angle correlation function finds all neighbor points whose angular distance are close and aggregate the features on the kernel points.}
    \label{fig:kernel}
\end{figure}

\noindent\textbf{Overview}  We proposed a local feature encoding module named Angle Correlation Point Convolution (ACPConv) to learn local shape and context. Given $N$ points in a point cloud $P \in \mathbb{R}^{N\times 3}$ and the corresponding features $F \in \mathbb{R}^{N\times C_{in}}$ carried by the points. For a point $p_i \in \mathbb{R}^3$ with feature $f_i \in \mathbb{R}^{C_{in}}$ in the point cloud, a generalized point convolution $\mathcal{G}$ maps the input features $\mathbb{R}^{N\times C_{in}}$ to the output features $\mathbb{R}^{N\times C_{out}}$:
\begin{equation}
    \mathcal{G} = \mathcal{F}_{p_j \in \mathcal{N}_j} (\mathcal{K}(p_i, p_j) f_j)
\label{eq:pconv}
\end{equation}
where $\mathcal{F}$ denotes the aggregation function in terms of \textit{max}, \textit{sum}, or \textit{mean} operations.  $\mathcal{K}$ refers to the kernel function for point pair $p_i$ and $p_j$ which learns the position mapping. $\mathcal{N}$ represents the neighbor points of $p_i$, $p_j \in \mathcal{N}_j, j\in M$ is one neighborhood point. To local the local neighborhood $\mathcal{N}_j$, we opt to use K-Nearest-Neighbor (KNN) algorithm to search for $M$ neighbor points. The reason is that empirically KNN is more scalable and efficient especially for large scale point cloud processing, and the performance gap between KNN and ball radius search is marginal.

\noindent\textbf{Kernel Initialization} Similar to KPConv \cite{thomas2019kpconv}, we use manually designed kernels to embed local shapes. Differ from KPConv which carefully hand-crafted the locations of kernel points, we seek simplicity by initializing the kernel points randomly. This random assignment of kernels is favored by our design of multi-scale local and global feature encoding, achieving more robust training and better performance in the experiments. Specifically, we initialize a total of $K$ kernel points in the 3D space, with one kernel point occupying the center (where the query point is located). To avoid the kernel points from being too far away from the neighbor points, we set a distance limit to contain the kernel points within the neighborhood.   

\begin{figure*}
    \centering
    \includegraphics[width=17cm]{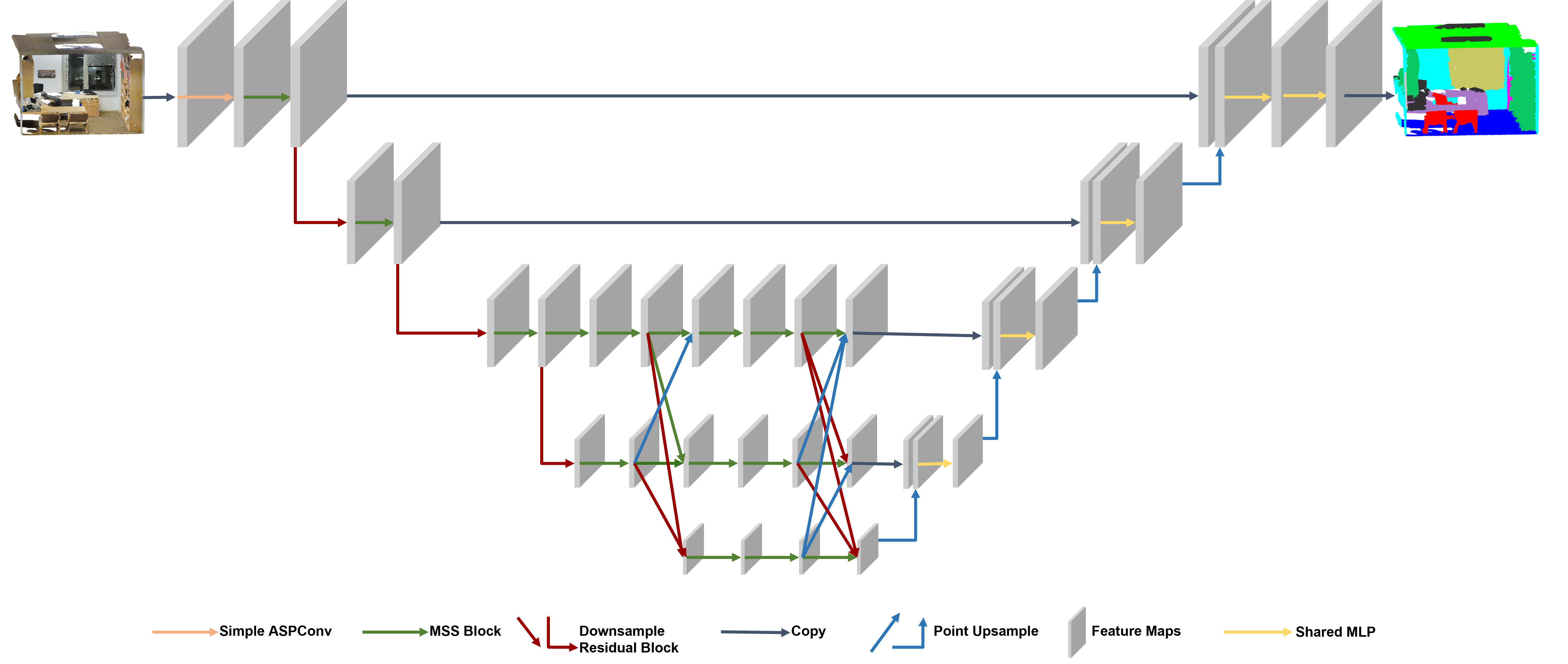}
    \caption{Overview of the proposed global multi-scale HRNet designed for point cloud semantic segmentation. High and low resolutions are hierarchically connected in the parallel streams. Through this information exchange, global representation is enhanced at multiple scales.}
    \label{fig:hrnet}
\end{figure*}

\begin{figure*}
    \centering
    \includegraphics[width=17cm]{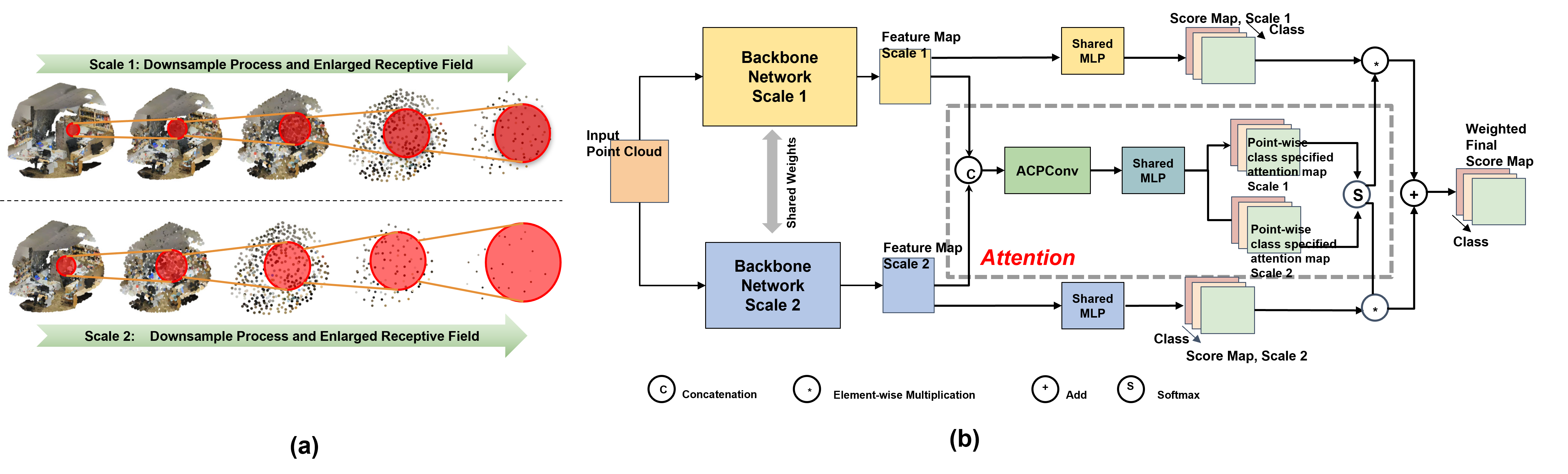}
    \caption{(a) Demonstration of the downsample process with different downsample rates.  (b) Illustration of the attentional multi-resolution fusion. The input point clouds are passed through the same network but with different downsampling rates. Attentional fusion (highlighted) is used to combine the predictions of both branches and achieve optimal results.}
    \label{fig:fusion}
\end{figure*}

\noindent\textbf{Angle Correlation}
With $K$ kernel points, our goal is to mimic 2D image convolution operation and aggregate the neighborhood features onto the kernel points. We use the angle-based function to define the correlation between neighbor points and kernel points (see Figure \ref{fig:kernel}). For a kernel point $p_k$, we find the neighbor points whose angular distance to $p_k$ is smaller than a threshold $\theta_t$. The correlation function between a neighborhood point $p_j$ and $p_k$ is defined as:
\begin{equation}
    Corr(p_j, p_k) = cos\; \theta (p_j, p_k)_{\theta < \theta_t} = \frac{p_j \cdot p_k}{\|p_j\| \; \|p_k\|}
\label{eq:corr}
\end{equation}
From Equation \ref{eq:corr}, we model the correlation as the cosine of the angular difference between $p_j$ and $p_k$, the smaller the angle, the bigger the correlation weight. 

Using the correlation function, our kernel function $\mathcal{K}$ in Equation \ref{eq:pconv} can be formulated as:
\begin{equation}
    \mathcal{K} = \sum_{k \in K} Corr(p_j, p_k) W_{k}
\label{eq:kernel}
\end{equation}
where $W_k \in \mathbb{R}^{C_{in}\times C_{out}}$ is the learnable weight matrix. Note that for the center kernel point which overlaps with the query point, it correlates with the query point only, $Corr$ is set to 1. 

\subsection{Local Multi-scale Split (MSS) Block}

In order to effectively encode the local shape and learn the context, we adopt a hierarchical multi-scale split block using our proposed ACPConv as the major component. Illustrated in Figure \ref{fig:hsblock}, the MSS block, inspired by Yuan et al. \cite{yuan2020hs}, learns the residual connection following the spirit of ResNet \cite{he2016deep}. 
After the shared MLP, we obtain a feature map with $d'$ channels. We equivalently split this feature map into four sub feature maps $x_i, i \in \{1,2,3,4\}$. We directly copy $x_1$ to the output feature map. Next, we feed $x_2$ into an ACPConv module. The output of this ACPConv is split evenly into two groups $y_{21}, y_{22}$. The first group $y_{21}$ is copied and concatenated with $x_1$ in the final output feature map. The second group $y_{22}$ is concatenated with $x_3$ and used as input to another ACPConv module, the output of which is split evenly into $y_{31}, y_{32}$. $y_{31}$ is concatenated with $y_{21}$ in the final output, while $y_{32}$ is concatenated with $x_4$ and passed through the third ACPConv. The output $y_4$ is concatenated with $x_1$, $y_{21}$, $y_{31}$. The MSS block gradually increases the receptive field, enabling the capturing of local multi-scale shapes and contexts.

\subsection{Global Multi-scale HRNet Architecture}
The network architectures vary across recent point cloud networks. A great number of works are the variants of MLP-based structure \cite{qi2017pointnet}, or they follow the paradigm of Unet-like network. In this work, we aim to construct a multi-level architecture that maintains a high-resolution representation. Therefore, we use a high-resolution network (HRNet) \cite{WangSCJDZLMTWLX19,SunXLW19} that has been successful for 2D image learning tasks. For the point cloud semantic segmentation tasks, high-resolution representation avoids semantic information loss and boosts the learning of multi-level contextual features. A detailed illustration of the proposed point cloud HRNet is shown in Figure \ref{fig:hrnet}. We simplify the original HRNet \cite{WangSCJDZLMTWLX19,SunXLW19} architecture for efficiency. For an accurate per-point semantic label prediction, we append a decoder that utilizes point upsampling to the HRNet backbone. Skip connection is used for recovering sharp low-level details. 
With the multi-resolution HRNet, we are able to learn global multi-scale information that is beneficial for semantic segmentation performance. 

In our implementation, we use Poisson disk sampling (PDS) to downsample points, and feature propagation as used in PointNet++ \cite{qi2017pointnet++} to upsample points. The reason we use PDS is that it is easy to manipulate the point uniformity and downsample resolution with Poisson disk sampling radius. 
In detail, we use the parameter radius $r$ for each level of downsampling. For each stage $l$ of downsampling, we double the radius $r_l = r * 2^{l-1}$.

\begin{table*}[hbt!]
\setlength\tabcolsep{0.5pt}
 \caption{Semantic segmentation quantitative comparisons on S3DIS \cite{armeni2017joint}, tested on Area 5. We reported mean intersection over union (mIou) (\%) and overall accuracy (\%) scores as well as mIoU for individual classes.}
\begin{scriptsize}
\begin{center}
\begin{tabular}{m{10em} | m{3.2em} | m{3.2em} | *{13}{m{3.2em}}}
\hline 
 Method	 & mIoU & OA & ceil.	 & floor	 & wall	 & beam	 & col.	 & wind.	 & door	 & chair	 & table	 & book.	 & sofa	 & board & clut.	\\
 \hline
PointNet \cite{qi2017pointnet}	& 41.1	& 49.0	& 88.8	& 97.3	& 69.8	& 0.1	& 3.9	& 46.3	& 10.8	& 52.6	& 58.9	& 40.3	& 5.9	& 26.4	& 33.2	\\
KPConv (rigid) \cite{thomas2019kpconv} & 65.4	& -	& 92.6	& 97.3	& 81.4	& 0.0	& 16.5	& 54.5	& 69.5	& 90.1	& 80.2	& 74.6	& 66.4	& 63.7	& 58.1	\\
KPConv (deform) \cite{thomas2019kpconv}  & 67.1	& - 	& 92.8	& 97.3	& 82.4	& 0.0	& 23.9	& 58.0	& 69.0	& 91.0	& 81.5	& 75.3	& 75.4	& 66.7	& 58.9	\\
SPH-GCN \cite{lei2020spherical}   & 59.5 & 87.7 & 93.3 & 97.1 & 81.1 & 0.0 & 33.2 & 45.8 & 43.8 & 79.7 & 86.9 & 33.2 & 71.5 & 54.1 & 53.7 \\
SPNet \cite{li2021spnet}	& 69.9	& 90.3	& 94.5	& 98.3	& 84.0	& 0.0	& 24.0	& 59.7	& 79.8	& 89.6	& 81.0	& 75.2	& 82.4	& 80.4	& 60.4	\\

 \hline
 \textbf{Ours } &\textbf{70.7} & \textbf{91.0} & \textbf{94.7}	& 98.2	& \textbf{86.2} & 0.0 & \textbf{45.8}	& \textbf{61.4} & 71.1 &	82.5 & \textbf{90.3} &	73.4 &	76.1 &	77.8 &	\textbf{61.2}\\
 \hline 
 \end{tabular}
 \end{center}
 \end{scriptsize}
 \label{tab:s3d} 
 \end{table*}
 
\begin{table*}[hbt!]
\setlength\tabcolsep{0.5pt}
 \caption{Semantic segmentation quantitative comparisons on ScanNet online test set \cite{dai2017scannet}. We reported mIoU (\%) scores.}
\begin{scriptsize}
\begin{center}
\begin{tabular}{m{9em} | m{3em} | *{20}{m{2.8em}}}
\hline 
 Method	 & mIoU & floor & wall & chair & sofa & table & door & cab &  bed & desk & toil & sink & wind & pic & bkshf & curt & show & cntr & fridg & bath & other	\\
 \hline
PointNet++ \cite{qi2017pointnet++}	& 33.9 & 67.7 & 52.3 & 36.0 & 34.6 & 23.2 & 26.1 & 25.6 & 47.8 & 27.8 & 54.8 & 36.4 & 25.2 & 11.7 & 45.8 & 24.7 & 14.5 & 25.0 & 21.2 & 58.4 & 18.3	\\
RandLa-Net \cite{hu2020randla} & 64.5 & 94.5 & 79.2 & 82.9 & 73.8 & 59.9 & 52.3 & 57.7 & 73.1 & 47.7 & 82.7 & 61.8 & 62.1 & 26.9 & 69.9 & 73.6 & 74.9 & 44.6 & 48.4 & 77.8 & 45.4 \\
SPH-GCN \cite{lei2020spherical} & 61.0 & 93.5 & 77.3 & 79.2 & 70.5 & 54.9 & 50.7 & 53.2 & 77.2 & 57.0 & 85.9 & 60.2 & 53.4 & 4.6 & 48.9 & 64.3 & 70.2 & 40.4 & 51.0 & 85.8 & 41.4 \\
KPConv (rigid) \cite{thomas2019kpconv}  & 68.4 & 93.5 & 81.9 & 81.4 & 78.5 & 61.4 & 59.4 & 64.7 & 75.8 & 60.5 & 88.2 & 69.0 & 63.2 & 18.1 & 78.4 & 77.2 & 80.5 & 47.3 & 58.7 & 84.7 & 45.0	\\
MinkowskiNet \cite{choy20194d} & 73.6 & 95.1 & 85.2 & 84.0 & 77.2 & 68.3 & 64.3 & 70.9 & 81.8 & 66.0 & 87.4 & 67.5 & 72.7 & 28.6 & 83.2 & 85.3 & 89.3 & 52.1 & 73.1 & 85.9 & 54.4 \\
VMVF \cite{kundu2020virtual} & 74.6 & 94.8 & 86.6 & 86.5 & 79.6 & 70.4 & 66.4 & 70.2 & 81.9 & 69.9 & 93.5 & 76.4 & 72.8 & 33.0 & 84.8 & 89.9 & 85.1 & 39.7 & 74.6 & 77.1 & 58.8\\ 
\hline
\textbf{Ours} & \textbf{69.6} & 94.1 & 84.0 & 84.1 & \textbf{80.6} & 60.2 & 59.1 & 67.5 & 75.5 & 56.6 & 88.6 & 68.7 & 66.5 & 31.0 & 81.4 & 81.3 & 74.7 & 49.8 & 57.0 & \textbf{86.4} & 45.0\\
 \hline 
 \end{tabular}
 \end{center}
 \end{scriptsize}
 \label{tab:scannet} 
 \end{table*}

\subsection{Multi-resolution Network with Attentional Fusion}
To further enhance the learning of multi-scale information, we propose a fusion mechanism that processes and fuses two different resolution predictions. For semantic segmentation tasks, some types of classes are best handled at a low resolution while other types are better handled at high resolution. For example, fine details such as the edges of objects are often more accurately predicted with high-resolution point clouds. Meanwhile, large building structures, which require more global context, are often done better at low-resolution, which observe more of the necessary context given the same receptive field of a network (an ablation can be found in Section \ref{sec:ablation}). Motivated by this, we design a point-wise multi-scale attention mechanism to adaptively fuse the multi-scale predictions. Shown in Figure \ref{fig:fusion}, we use two different downsampling rates $r_1$ and $r_2 = 2*r_1$  for two parallel branches. Therefore, the receptive fields of the two branches vary, capturing different levels of information. To effectively fuse the multi-scale outputs,
we concatenate the feature maps before the prediction layer and use an ACPConv which outputs a point-wise attention score map for individual classes. The attention scores are used to re-weight probability maps of two branches. After re-weighting, the probabilities are added to produce the final prediction.

\section{Experiments}

\subsection{Datasets}
For semantic segmentation, we use S3DIS \cite{armeni2017joint} and ScanNet \cite{dai2017scannet} for indoor scenes, and NPM3D \cite{roynard2018paris} for outdoor scenes.
S3DIS \cite{armeni2017joint} contains 3D scans from Matterport scanners in 6 indoor areas including 271 rooms. Each point is annotated with
one of 13 categories of semantic labels. We use Area 5 for testing and use other areas for training.  ScanNet \cite{dai2017scannet} consists of 1513 RGB-D reconstructed small indoor scenes annotated in 20 categories. We use 1201 scenes for the training set, 312 scenes for validation, and 100 scenes (no ground truth label available) for testing.
NPM3D (Paris-Lille 3D) \cite{roynard2018paris} contains more than $2km$ of streets
in 4 different cities. The 160 million points of this dataset are annotated with 10 semantic classes. 

\begin{table}[hbt!]
 \caption{Segmentation quantitative comparisons on S3DIS \cite{armeni2017joint}, 6-fold and NPM3D \cite{roynard2018paris}. We reported mIou (\%) Scores.}
\begin{center}
\begin{tabular}{m{10em} | m{6em} |  m{4em}}
\hline
Method & S3DIS 6-fold (mIoU) & NPM3D (mIoU)\\
\hline
KPConv \cite{thomas2019kpconv} & 69.6 & \textbf{82.0}\\
ShellNet \cite{zhang2019shellnet} & 66.8 & -\\
SPH-GCN \cite{lei2020spherical} & 68.9 & -\\
RandLa-Net \cite{hu2020randla} & 70.0 & 78.5\\
\hline
\textbf{Ours} & \textbf{73.5} & 80.1\\
\hline
\end{tabular}
 \end{center}
 \label{tab:npm3d} 
 \end{table}
 
 \begin{figure}
    \centering
    \includegraphics[width=7cm]{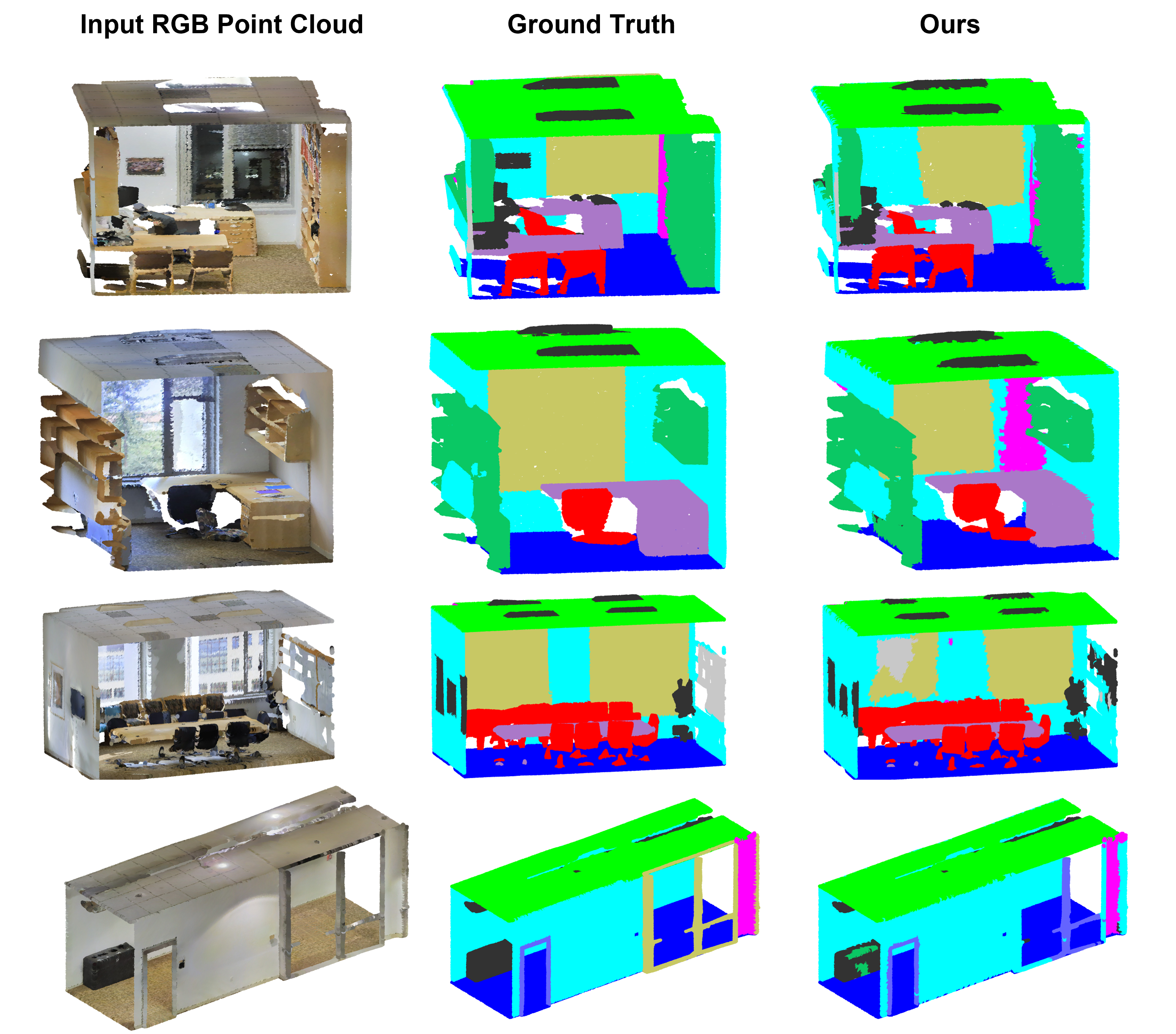}
    \caption{Visualization of S3DIS segmentation results.}
    \label{fig:qualitative}
\end{figure}

\begin{table*}[hbt!]
 \caption{Abaltion study of individual components of our method, evaluated on S3DIS \cite{armeni2017joint} Area 5. }
\begin{center}
\begin{tabular}{m{1em} m{30em} | m{4em} m{4em} m{4em}}
\hline
& Method & mIoU & Gain & \#params\\
\hline
(1) & Unet + ACPConv + ResNet block &  68.2 & - & 14.1M\\
(2) & Unet + ACPConv + MSS block &  69.1 & +0.9 & 10.9M\\
(3) & HRNet + ACPConv + ResNet block  & 69.6 & +1.3 & 19.5M\\
(4) & HRNet + ACPConv + MSS block  & 70.3 & +2.1 & 16.6M\\
(5) & HRNet + ACPConv + MSS block + Attentional Fusion (full pipeline) & \textbf{70.7} & +2.5 & 17.0M\\

\hline 
\end{tabular}
 \end{center}
 \label{tab:ablation} 
 \end{table*}

\begin{table*}[hbt!]
\setlength\tabcolsep{0.5pt}
 \caption{Ablation Study on multi-resolution fusion, evaluted on S3DIS \cite{armeni2017joint} Area 5.}
\begin{scriptsize}
\begin{center}
\begin{tabular}{m{2em} m{17em} | m{3.2em} | *{13}{m{3.2em}}}
\hline 
  &	 & mIoU  & ceil.	 & floor	 & wall	 & beam	 & col.	 & wind.	 & door	 & chair	 & table	 & book.	 & sofa	 & board & clut.	\\
 \hline

(1) & Scale 1, downsample rate = 0.04m & 68.2	& 92.9	& 98.1	& 83.7	& 0.0 & 27.7 & 71.7	& 57.4	& 81.7	& 90.9	& 73.4	& \textbf{76.8}	& 72.5	& 60.1	\\

(2) & Scale 2, downsample rate = 0.08m & 67.6	& 94.7	& 98.3	& \textbf{85.8}	& 0.0 & 40.2 & \textbf{75.7}	& \textbf{58.8}	& 82.3	& 89.0	& 48.6	& 74.0	& 71.3	& 60.2	\\
(3) & Fusion of scale 1 and scale 2, average & 69.6 & \textbf{95.4} & 	98.1 & 83.5	& 0.0 & 37.7 & 	74.5	 & 57.3	 & 83.1	 & 91.1	 & 74.7	 & 76.2	 & 70.8	 & \textbf{62.8	}\\
(4) & Fusion of scale 1 and scale 2, attention & \textbf{70.0}	 & 94.9	 & \textbf{98.3}	 & 83.8	 & 0.0	 & \textbf{41.4}	 & 75.0	 & 54.6	 & \textbf{83.4}	 & \textbf{91.6}	 & \textbf{74.8}	 & 76.4	 & \textbf{75.0}	 & 61.3	\\
 \hline 
 \end{tabular}
 \end{center}
 \end{scriptsize}
 \label{tab:multi-res} 
 \end{table*}
 
\subsection{Implementation Details}
Similar to the previous methods \cite{qi2017pointnet++,thomas2019kpconv}, we do not directly process the entire scene. 
During training, we randomly pick center locations in each scene and query for spheres with fixed radius. During testing, to ensure all points are predicted at least once, we uniformly pick the center locations. The sphere radius are $2.5m$ for S3DIS \cite{armeni2017joint} and ScanNet \cite{dai2017scannet}, and $5m$ for NPM3D. We adopt the same voting scheme as \cite{thomas2019kpconv}, the probabilities for each point are averaged. For indoor datasets where color is available, we use RGB as input features in addition to XYZ coordinates. A constant 1 feature is used for all datasets to ensure black points are not ignored. The augmentation techniques including rotation, coordinates perturbation, scaling, and color dropping (only applied for S3DIS).  
We set ACPConv angle correlation function threshold $\theta_t$ to $30^{\circ}$. We use $K=15$ kernel points, and $M=32$ number of neighbor points.
For multi-resolution fusion,  we use downsample rates to $r_1=0.04m, r_2=0.08m$ for indoor datasets, and $r_1=0.08m, r_2=0.12m$ for outdoor datasets. We use Adam optimizer \cite{kingma2014adam} with default setting for all experiments. The initial learning rate is set to 0.001, and decays by $50\%$ every 30 epochs. We train S3DIS and ScanNet for 120 epochs and NPM3D for 80 epochs. Cross-entropy is used as the loss function.

\subsection{Overall Performance}
The quantitative performance comparisons including individual class scores of S3DIS (Area 5) \cite{armeni2017joint} between other methods and ours are shown in Table \ref{tab:s3d}. The results of ScanNet \cite{dai2017scannet} are shown in Table \ref{tab:scannet}. 
We use mean intersection over union (mIoU) and overall accuracy (OA) as evaluation metrics. From Table \ref{tab:s3d} \ref{tab:scannet} \ref{tab:npm3d}, our method achieves significant performances. Note that our method dominates in several classes that are difficult to segment. For example, we achieve the best score on column (S3DIS), suggesting our method is able to learn building structures. We also perform well on picture and curtain (ScanNet), suggesting our method captures the details of objects. 
In Figure \ref{fig:qualitative}, We show qualitative results of our method on several test scenes on S3DIS.

\subsection{Ablation Studies}
\label{sec:ablation}
 \noindent\textbf{Framework Configuration}
We conduct ablation experiments to investigate design variations and demonstrate the
advantages of our method. 
In Table \ref{tab:ablation}, we show the ablation studies by adding one component at a time. Our baseline method (Table \ref{tab:ablation} (1)) is formulated as a Unet in which regular ResNet-like bottleneck block \cite{he2016deep} and ACPConv are used. The Unet has 4 stages of downsampling and upsampling, similar to the network used in KPConv \cite{thomas2019kpconv}.
Notably, our method has a 67.2\% mIoU score that is on par with current state-of-the-arts. After replacing the ResNet block with our MSS block (Table \ref{tab:ablation} (2)), we see a 1.9\% performance improvement and 3.2M parameter reduction, demonstrating the effectiveness and efficiency of the MSS block. While keeping the same ResNet block and replacing Unet with HRNet (3), we see a 1.3\% improvement compared to the baseline (1). Using MSS block and HRNet architecture (4), the performance increases 1.1\%, demonstrating the success of our local and global multi-scale representation. To show the effectiveness of the multi-scale fusion, we keep the same configuration as used in (4), and use the attentional fusion of two resolutions (5). The performance gain is 2.5\% compared to the baseline. 

\noindent\textbf{Attentional Fusion} 
Table \ref{tab:multi-res} shows the ablation for the multi-resolution fusion. For fair comparison, we use the baseline configuration (Table \ref{tab:ablation} (1)) for these experiments.
We use two different sampling rates and report the performances (Table \ref{tab:multi-res} (1) and (2)). Scale 1 has a small downsample rate, therefore fewer points are discarded through the feature encoding, performing better in classes such as bookshelf and sofa. Scale 2 on the other hand, performs better in classes such as column and window. This phenomenon shows that fine detail, such as furniture, is often better predicted with small downsample rates. Predictions of large structures, which require more global context, are often done better at a larger downsample rate. We conduct two experiments to fuse the multi-resolution predictions. In (3), we simply average the probabilities of two branches. The score of (3) is higher than both (1) and (2), suggesting this simple fusion method already has a weak capability of generating better predictions over multi-resolutions. The proposed attentional fusion (4) shows the highest score over all the experiments. Experiment (4) achieves the best performances on both furniture classes (such as bookshelf) and building structure classes (such as column).

\section{Conclusion}
In this paper, we introduce a point cloud semantic segmentation framework that effectively learns local and global multi-scale information. Our proposed multi-scale representation is beneficial for learning point cloud shapes and context. Moreover, to combine the predictions from multi-resolution inputs, we propose an attention fusion mechanism. Our experimental results show that our method achieves top-ranking performances in several benchmarks. Our extensive ablations demonstrate the effectiveness of individual modules.






%
\clearpage
\bibliographystyle{IEEEtran}
\bibliography{egbib}

\end{document}